\documentclass[letterpaper]{article} 
\usepackage{aaai23}  
\usepackage{times}  
\usepackage{helvet}  
\usepackage{courier}  
\usepackage[hyphens]{url} 
\usepackage{balance} 
\usepackage{graphicx} 
\usepackage{natbib}  
\usepackage{caption} 
\usepackage{algorithm}
\usepackage{algorithmic}

\usepackage{newfloat}
\usepackage{listings}
\DeclareCaptionStyle{ruled}{labelfont=normalfont,labelsep=colon,strut=off} 
\floatstyle{ruled}
\newfloat{listing}{tb}{lst}{}
\floatname{listing}{Listing}
\usepackage{subfig}
\usepackage{multirow}
\usepackage{comment}

\def\eg{\emph{e.g., }}
\newcommand{\bfsection}[1]{\vspace*{0.1cm}\noindent\textbf{#1.}}

\title{A New Dataset and Comparative Study for Aphid Cluster Detection}
\author{
    Tianxiao Zhang\textsuperscript{\rm 1},
    Kaidong Li\textsuperscript{\rm 1},
    Xiangyu Chen\textsuperscript{\rm 1},
    Cuncong Zhong\textsuperscript{\rm 1},
    Bo Luo\textsuperscript{\rm 1},
    Ivan Grijalva Teran\textsuperscript{\rm 2},
    Brian McCornack\textsuperscript{\rm 2},
    Daniel Flippo\textsuperscript{\rm 3},
    Ajay Sharda\textsuperscript{\rm 3},
    Guanghui Wang*\textsuperscript{\rm 4}
}
\affiliations{
    \textsuperscript{\rm 1}Department of Electrical Engineering and Computer Science, University of Kansas, Lawrence, KS 66045, USA\\
    \textsuperscript{\rm 2}Department of Entomology, Kansas State University, Manhattan, KS 66506, USA\\
        \textsuperscript{\rm 3}Department of Biological and Agricultural Engineering, Kansas State University, Manhattan, KS 66506, USA\\
    \textsuperscript{\rm 4}Department of Computer Science, Toronto Metropolitan University, Toronto, ON M5B 2K3, Canada\\

}

\usepackage{bibentry}

\begin{document}

\maketitle

\begin{abstract}
Aphids are one of the main threats to crops, rural families, and global food security. Chemical pest control is a necessary component of crop production for maximizing yields, however, it is unnecessary to apply the chemical approaches to the entire fields in consideration of the environmental pollution and the cost. Thus, accurately localizing the aphid and estimating the infestation level is crucial to the precise local application of pesticides. Aphid detection is very challenging as each individual aphid is really small and all aphids are crowded together as clusters. In this paper, we propose to estimate the infection level by detecting aphid clusters. We have taken millions of images in the sorghum fields, manually selected 5,447 images that contain aphids, and annotated each aphid cluster in the image. To use these images for machine learning models, we crop the images into patches and created a labeled dataset with over 151,000 image patches. Then, we implement and compare the performance of four state-of-the-art object detection models. 
\end{abstract}

\section{Introduction}

Annually 37\% of crops are lost to pest damage and around 13\% of crop damage is caused by insects. Most farmers consider utilizing pesticides to eliminate insects and a tremendous amount of funding were applied to pesticides each year. While most of the pesticides are wasteful since only a small portion of used pesticides is employed directly on the insects and most of them are wasted and even pollute the environment. 
Under several management scenarios, only a small fraction of areas receive a justified amount of pesticide, while other areas lose yield due to delayed timing and damage by pests, and remaining areas receive a superfluous spray application when there is no pest presence. 
However, the development of robotic technology for insecticide application has not been explored primarily due to unavailable camera vision capabilities to locate the pest incidence and severity within a complicated crop canopy. There is an urgent demand for an intelligent application system designed to accurately spray on the infested canopy but only where infestations are present. 

Object detection and recognition is one of the most critical components in agricultural robotics, and detecting small insects like aphids can be especially challenging. Convolutional neural network (CNN) was first used in \cite{girshick2014rich} for object detection and recognition. 
The CNN models have a wide range of applications on medical image analysis \cite{li2021colonoscopy} and object detection \cite{ma2021miti}\cite{zhang2020real}. 
Nonetheless, aphids are so tiny that even state-of-the-art detection models could not accurately localize them individually. 
Most existing aphid detection models \cite{teng2022td}\cite{li2019automatic}\cite{li2019coarse} attempt to detect individual aphids, but the performance of the detection models on the aphid dataset is still not that pleasant. In addition, those models are trained on ideal aphid images, it is even harder for them to detect the aphid in real-world scenarios since most aphids are clustered together on the leaves, and accurately dividing and detecting the dense aphids individually is almost impossible. In addition, different illuminations and shades in different images might cause domain shifts which also severely affect the accuracy of CNN models on the tiny aphids detection \cite{yang2022unsupervised}\cite{zhang2021six}.

\begin{figure}[t]
\begin{center}
   \includegraphics[width=1.0\linewidth]{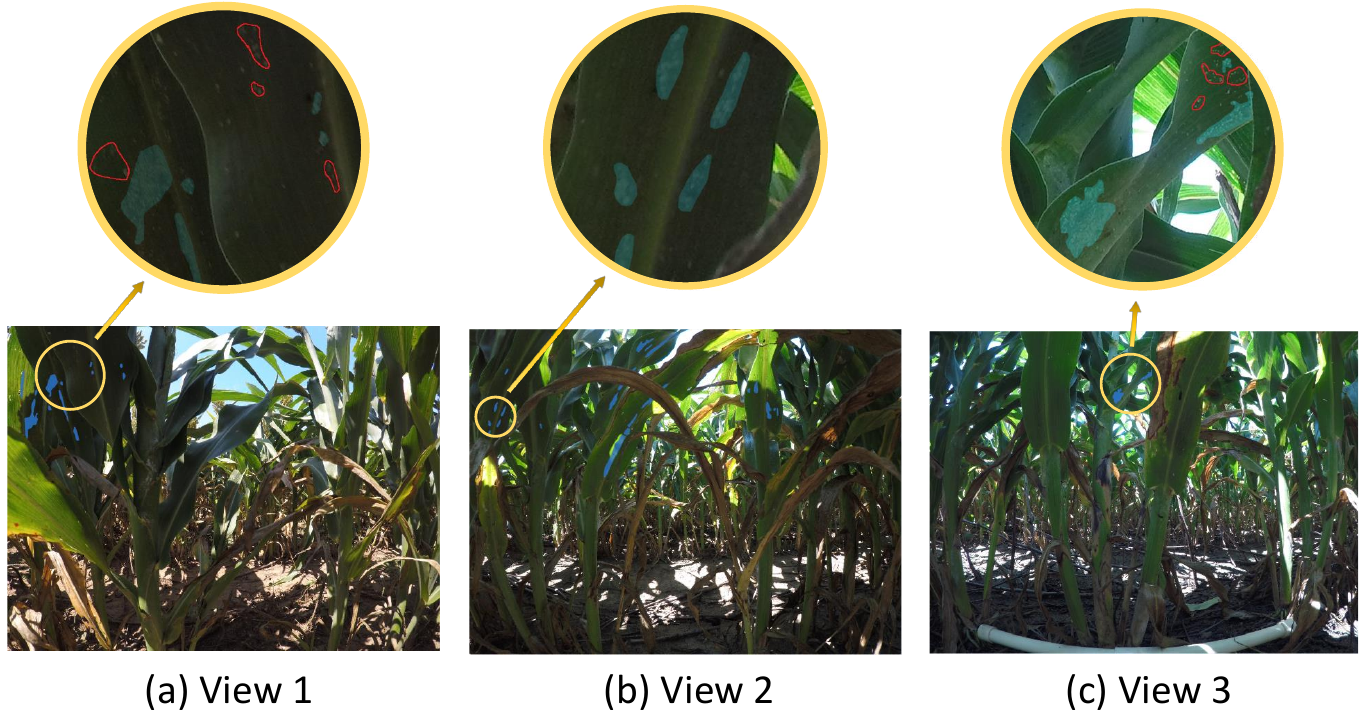}
\end{center}
   \caption{{Original images with annotations}. The light blue areas represent an annotated aphid cluster. Aphid clusters are mostly tiny compared to the original image size. In (a) and (c), the areas circled by red lines represent clusters we do not need to annotate since those areas only contain very few sparsely distributed aphids. The criteria is an aphid cluster should have more than 6 closely located aphids.}
\label{fig:samples}
\end{figure}

\begin{figure*}[t]
\centering
    \subfloat{
        \includegraphics[height=4.8cm]{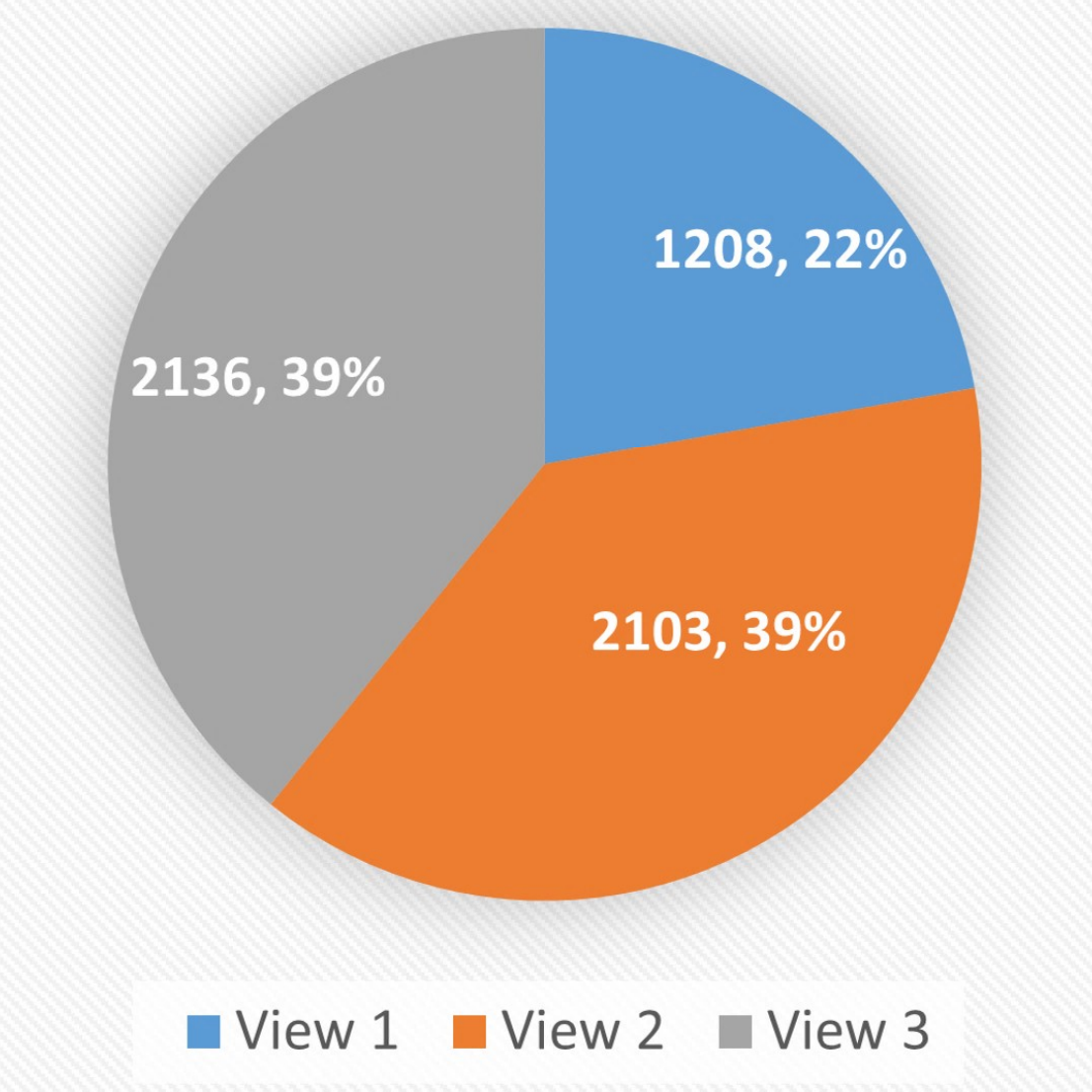}
        \label{plots:compo}
    }
    \hspace{12pt}
    \subfloat{
        \includegraphics[height=4.8cm]{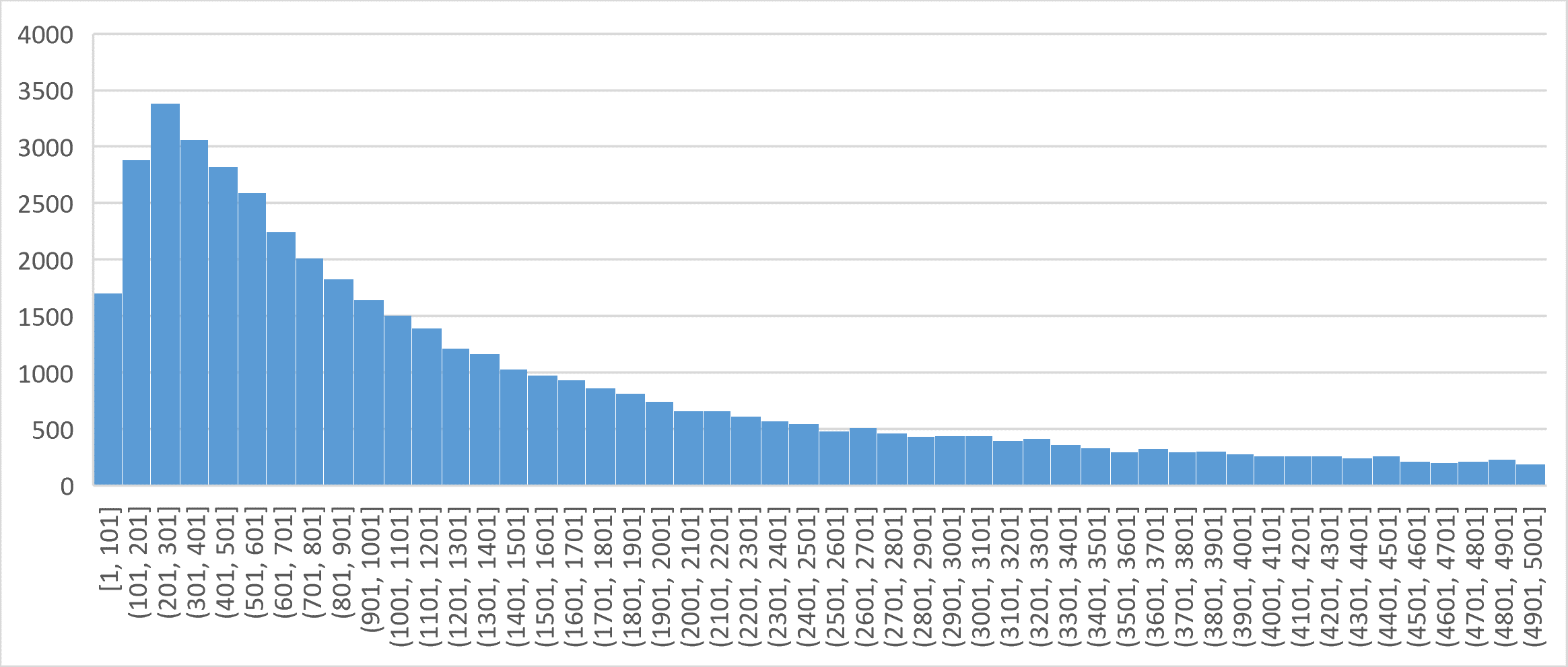} 
        \label{plots:distr}
    }
	\caption{{Statistical summary of the created dataset. (left) Dataset composition; (right) mask size distribution}.}
	\label{fig:set_sum}
\label{fig:s2}
\end{figure*}

In this paper, we collected millions of images from the sorghum field over two seasons, and then manually selected and annotated 5,447 images affected by aphids. Instead of labeling the aphid individually, we propose annotating aphid colonies as clusters and creating the bounding boxes based on the aphid clusters. Because bounding boxes are always rectangles and cover larger areas, closely located bounding boxes are merged together without affecting aphid detection. 
In addition, we implemented and evaluated the performance of four state-of-the-art detection models on the generated aphid dataset. 
The study makes it possible to estimate the aphid affection levels from real images to assist farmers to make timely control of affection. The labeled dataset and developed learning models will be freely accessible to the research community on the author's homepage. 

\section{Dataset Generation}



\subsection{Data Collection}

Most of the aphids are located under the leaves and majority of them are densely clustered. 
In order to reduce the influence of occlusion among the sorghum leaves, we developed an imaging rig with three GoPro cameras that can capture the canopy leaves at three different heights corresponding to view 1, view 2, and view 3, respectively. Thus, we can enrich the dataset by taking pictures at three different views and capturing aphid clusters from different perspectives. Three sample annotated images corresponding to the three views are shown in Fig~\ref{fig:samples}. Using this device, we have captured millions of images over two growing seasons of a sorghum farm. Most of the images are free of aphids. We manually examine all images and eliminate those without aphids, and eventually, select 5,447 images that contain adequate aphids. The percentages of photos corresponding to the three views are shown in Fig~\ref{fig:s2}.

\subsection{Data Labeling}
The aphid clusters in the selected images are manually annotated by professionally trained researchers using Labelbox\footnote{https://labelbox.com/}. 
We first create segmentation masks for each image and then generate detection bounding boxes based on the masks. In total, we have labeled 59,767 aphid clusters. 

\bfsection{Aphid Cluster Definition} Data labeling is a labor-intensive process. The task is distributed among 8 trained research assistants. Therefore, it is crucial to have an efficient and consistent definition of what is an aphid cluster before labeling. Ambiguous criteria will confuse deep learning models during training. 

In the fields, the aphid clusters can appear in a variety of patterns (low density, high density, different sparsity) as shown in Fig~\ref{fig:samples}. If we aim to label each individual aphid regardless of its density, it will take an excessive amount of time and resources will be wasted on areas without critical threat. If the threshold is set too high, areas with substantial aphid infestation might be ignored, resulting in financial loss.  After discussions with agricultural experts, we define the aphid cluster as ``an area with more than or equal to six closely located aphids". A further interpretation of the threshold is demonstrated in Fig~\ref{fig:samples}.
\begin{table*}[t]
\centering
\caption{{Number of patches in each group}}
\begin{tabular}{ | c | c | c | c | c | c | c | c | c | c | c | }
    \hline
	Sum  &  1  &  2  &  3  &  4  &  5  &  6  &  7  &  8  &  9  &  10 \\
    \hline
	151,380 & 14,778 & 15,392 & 14,567 & 15,720 & 15,943 & 14,929 & 15,272 & 15,276 & 14,140 & 15,363 \\
    \hline
\end{tabular}
\label{table:num_patch}
\end{table*}

\bfsection{Labeling}
After removing some redundant images and images without aphid clusters, we have labeled in total 5,447 photos. Redundant images are those taken from very close viewpoints, resulting in visually similar images. Photos without clusters are removed because we believe deep CNN models can learn sufficient negative features from empty spaces of the remaining photos. 

In summary, the statistical information of the generated masks is shown in Fig~\ref{fig:s2}. In total, 59,767 masks are created and the sizes of masks vary greatly. 77.0\% of the masks have a size smaller than 5,000 pixels. Since masks with larger sizes are rare and sparsely distributed, we only plot the histogram of masks with less than 5,000 pixels in Fig~\ref{plots:distr}. More than half of the masks are smaller than 1,500 pixels, with the most popular size interval [201, 301]. Among all masks, the median size is 1,442 pixels and the mean is 7,867 pixels. The median is more representative, while the mean value is severely affected by the extremely large masks.

\bfsection{10-Fold Cross Validation}
Cross validation \cite{stone1974cross} is a resampling method to evaluate and pick models on a small dataset. Popular computer vision datasets commonly have more than $10k$ images, \eg MS COCO \cite{lin2014microsoft} has more than $200k$ labeled images. Our dataset only has a little over $5k$ images. Following cross validation \cite{stone1974cross}, we decide to split our dataset into 10 groups. To ensure each group has a similar percentage of images from the three different views, we separately shuffle the images and split them into 10 subgroups from each view. Then the final cross validation groups are formed by picking one subgroup from each view. Thus images from each view will be evenly distributed in each group. 

\bfsection{Image Patches}
The majority of the masks, as shown in Fig~\ref{fig:s2}, have a size smaller than 1,500 pixels, which is less than 0.015\% of the original image size ($3,648 \times 2,736$). In addition, most detection and segmentation models are trained and tested on much smaller images. So we crop the original high-definition images into smaller $400 \time 400$ patches. During this process, some masks will be separated into different patches and will have some exclusions. To ensure each mask's completeness in at least one of the final patches, the patch generation is performed with 50\% overlapping, meaning the next patch overlaps 50\% with the previous patch both horizontally and vertically. An original $3,648 \times 2,736$ image will generate 221 patches for detection and segmentation. 

Patch generation is conducted after dividing the dataset into 10 cross validation groups, such that information from one original photo will not leak to any other groups. Also after patch generation, those patches without an aphid cluster are discarded because CNN models should have enough negative samples just from the background of other patches. In summary, the number of patches in each cross validation group is shown in Table \ref{table:num_patch}.

\begin{figure}[t]
\begin{center}
   \includegraphics[width=1.0\linewidth]{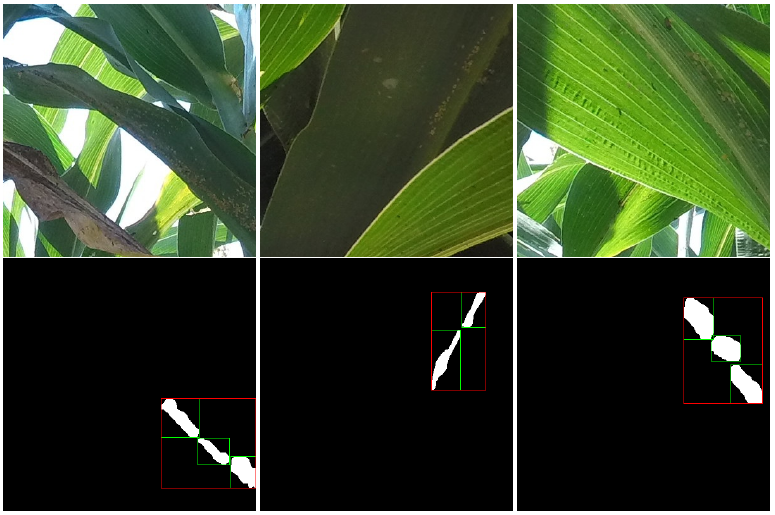}
\end{center}
   \caption{Some examples of merging clusters that are close enough (10 pixels) to each other. The segmentation masks are utilized to illustrate the changes in the bounding boxes before and after the merging process. The white areas denote the masks for aphid clusters. The green bounding boxes in the masks are the originally labeled boxes and the red bounding boxes in the masks are the merged results. It is very challenge for the detection models to detect individual cluster in thess scenarios. Thus, they are merged to form a more meaningful cluster.}
\label{fig:2}
\end{figure}

\bfsection{Bounding Box Merge} Since we label the aphids based on clusters, small clusters close to each other are labeled individually with well-defined boundaries. However, the generated bounding boxes of these clusters overlap with each other, as shown in Figure~\ref{fig:2}. From the object detection point of view, these bounding boxes should be merged as they all represent aphid clusters. Otherwise, they may cause confusion during learning. In our application, we merge the bounding boxes of the clusters if their closest distance is less than or equal to 10 pixels. Our experiments show that this process will greatly boost detection accuracy.

\bfsection{Tiny Cluster Removal} The process of image cropping may create some extremely small clusters, and most of them are around the border or corner of the patches. In practice, these small labels are meaningless for model training and affection estimation. Thus, we remove the small cluster masks whose areas are less than 1\% of the patch. The results after merging and removal are illustrated in Table~\ref{table:1}.

\section{Object Detection Models}

In object detection, both classification and localization are required for recognizing and localizing the objects in the videos or images. Typically, the detection models have two separate branches for classification and localization, respectively. The classification branch is similar to most classification tasks which classify the contents included by the bounding boxes. The localization branch predicts the offsets to the anchor boxes for anchor-based detection models or to the anchor points for anchor-free detection models and then the offsets would be converted to the bounding box coordinates based on the anchor boxes or anchor points for final predictions. 
Since the IoU thresholds are extremely important for detection models, recent detection models tend to calculate the adaptive thresholds based on the statistical properties among the samples \cite{zhang2020bridging}\cite{zhang2021varifocalnet} or compute the dynamic thresholds based on the training status \cite{kim2020probabilistic}\cite{zhang2022dynamic}.

In this study, we implemented the following four state-of-the-art object detectors and evaluated their performance on aphid detection based on the created dataset. (1) ATSS (Adaptive Training Sample Selection) \cite{zhang2020bridging} calculates the adaptive IoU thresholds based on the mean and standard deviation of the IoUs between the candidate anchor boxes and the ground truth objects to select the positive samples instead of using fixed thresholds. (2) GFLV2 (Generalized Focal Loss V2) \cite{li2021generalized} utilizes statistics of bounding box distributions as the Localization Quality Estimation (LQE). Thus the high-quality bounding boxes could have a high probability to be kept instead of suppressed with the NMS (Non-Maximum Suppression) algorithm. (3) PAA \cite{kim2020probabilistic} dynamically divides the positive samples and negative samples using GMM (Gaussian Mixture Model) based on the classification and localization scores of the samples in a probabilistic way. (4) VFNet \cite{zhang2021varifocalnet} is based on ATSS \cite{zhang2020bridging} algorithm, but proposes IoU-aware Classification Score (IACS) as the classification soft target using the IoUs between the predicted bounding boxes and their corresponding ground truth objects. Thus high-quality predicted boundary boxes might have high scores than those low-quality boxes. In addition, star-shaped box feature representation is introduced to further refine the predicted boxes so that they could be closer to the ground truth objects.

The aforementioned detection models are state-of-the-art approaches that have excellent performance on COCO benchmark \cite{lin2014microsoft}. Since the labels of our created dataset are based on the aphid clusters instead of the single aphid, we can directly apply them in this problem and train these models using the created dataset.

\begin{table}[t]
\setlength{\tabcolsep}{4pt}
\small
\centering
\caption{The mean and standard deviation of 10-fold cross validation on state-of-the-art detection models}

\renewcommand{\arraystretch}{1.5}
\begin{tabular}{c | c | c | c | c}
\hline
        &&         original & +merge 10    &   +rm 0.01  \\
\hline
\multirow{2}{4em}{VFNet} & AP & $41.9\pm1.91$ & $44.8\pm1.89$  & $58.3\pm1.87$ \\
&recall& $80.4\pm0.90$ & $83.7\pm0.92$  & $96.8\pm0.35$ \\
\hline
\multirow{2}{4em}{GFLV2} & AP & $41.6\pm1.84$ & $44.7\pm1.93$  & $58.3\pm1.90$ \\
&recall& $79.2\pm1.27$ & $82.6\pm1.09$  & $96.2\pm0.45$ \\
\hline
\multirow{2}{4em}{PAA} & AP & $41.2\pm1.65$ & $44.2\pm1.68$  & $58.7\pm1.89$ \\
&recall& $84.1\pm0.89$ & $87.6\pm0.86$  & $98.4\pm0.23$ \\
\hline
\multirow{2}{4em}{ATSS} & AP & $41.8\pm1.70$ & $44.8\pm1.85$  & $59.0\pm1.82$ \\
&recall& $80.0\pm1.00$ & $83.3\pm1.03$  & $97.0\pm0.28$ \\
\hline

\end{tabular}
\label{table:1}
\end{table}

\subsection{Model Training}


All models exploit 0.001 as the initial learning rate with the total training epoch being 12. The initial learning rate is utilized for 9 epochs and then reduced by 10 times for the last 3 epochs. SGD (Stochastic Gradient Descent) is employed as the optimizer to optimize the model. The momentum and weight decay are 0.9 and 0.0005, respectively. The batch size is 16 and the warmup iterations are 500. The detection models are written by PyTorch with Python3 \cite{chen2019mmdetection}.
The evaluation metric for detection models is Average Precision (AP) which computes the area under the PR curve. The PR curve plots the Precision rate versus the Recall rate for the detection models. The precision rate indicates the correctly predicted samples over the entire predicted samples. The recall rate represents the correctly predicted samples over the entire ground truth samples. 



For object detection, only predicting the correct labels is not enough since we should also consider the bounding box accuracy. Typically, IoUs (Intersection over Unions) between the predicted bounding boxes and their corresponding ground truth boxes are utilized to judge the quality of the predicted boxes. Typically, IoU is calculated by the ratio of the intersection area over the union area of two bounding boxes. PASCAL VOC \cite{everingham2010pascal} selects 0.5 as the IoU threshold which indicates that the detection is a success if the IoU between the predicted bounding box and the ground truth bounding box is over 0.5 if the classification label is correctly predicted. COCO \cite{lin2014microsoft} chooses the IoU threshold from 0.5 to 0.95 with the step of 0.05, and calculates the AP for each of the thresholds and finally averages them. In this paper, we utilize the IoU threshold from PASCAL VOC and the generated annotation files are also in xml format as PASCAL VOC \cite{everingham2010pascal}.

\section{Results}

The performance across the 10-fold cross validation among all detection models is illustrated in Table~\ref{table:1}. AP (Average Precision) and Recall are recorded in the format of $mean\pm std$. The mean and standard deviation (std) are calculated across all 10-fold validations. In the first row, ``original" indicates the detection models are applied to the originally labeled dataset; ``+merge 10" illustrates the results after merging the clusters whose closest distance is within 10 pixels; ``+rm 0.01" stands for the results after removing the small clusters whose areas are less than 1\% of the patches. 

We can see from Table~\ref{table:1} that all detection models achieve similar results in terms of 
average precision, however, the recall rate of PAA \cite{kim2020probabilistic} is slightly higher than the other detectors since the number of predicted bounding boxes of PAA is higher than that of other detection models. Both AP and recall rate have been greatly increased after cluster merging (+merge 10) and removal (+rm 0.01). 
The above results are obtained with IoU setting to 0.5. We have also tested the influence of other IoU thresholds. In general, lower IoU yields better average accuracy and vice versa. If the locations of the aphid clusters are not concerned, a lower IoU threshold may be applied.

\section{Conclusion}

In this paper, we have selected thousands of aphid-affected images from millions of images captured in the fields and annotated the aphids based on clusters instead of individual aphids for generic usage. Due to the irregular shapes and sizes of the aphid clusters, the initial annotations might not be suitable for the training of learning models. Thus, we merge the bounding boxes of the neighboring clusters and remove the extremely small clusters. We have also evaluated and compared the performance of four state-of-the-art detectors and created a baseline of aphid detection using the created dataset. The experiments have also demonstrated the effectiveness of cluster merging and removal. The created dataset and trained detection models could be used to help farmers to estimated the aphid infestation levels in the field so as to provide timely and precise pesticide application. We hope our created dataset and analysis could inspire more work on aphid detection.

\section{Acknowledgments}
{This work was partly supported in part by NSERC (RGPIN-2021-04244) and USDA (2019-67021-28996).}

\balance
\bibliography{aaai23}

\end{document}